\documentclass[11pt,letterpaper]{article}
\usepackage[T1]{fontenc}
\usepackage[utf8]{inputenc}
\usepackage[margin=1in]{geometry}
\usepackage{times}
\usepackage{amsmath,amssymb,graphicx,booktabs,tabularx,longtable,array}
\usepackage[round,authoryear]{natbib}
\usepackage{authblk,needspace}
\usepackage{hyperref,url,placeins}
\hypersetup{hidelinks,pdftitle={OpenProblemBench: Benchmarking AI on Open Problems in the Foundational Theoretical Sciences},pdfauthor={Zhiyi Li, Sihan Hu, Tianning Xiao, Xiansheng Cai, Xiaojun Tan, Youjin Deng, Kun Chen}}
\newcommand{\bench}{OpenProblemBench}
\newcolumntype{Y}{>{\raggedright\arraybackslash}X}

\title{\Large\bfseries OpenProblemBench: Benchmarking AI on Open Problems in the Foundational Theoretical Sciences}
\author[1,2,3]{\mbox{Zhiyi Li}}
\author[1,2,3]{\mbox{Sihan Hu}}
\author[3,4]{\mbox{Tianning Xiao}}
\author[2]{\mbox{Xiansheng Cai}}
\author[5,6]{\mbox{Xiaojun Tan}}
\author[1,3,4]{\mbox{Youjin Deng}}
\author[2]{\mbox{Kun Chen}\thanks{Corresponding author. Email: \href{mailto:chenkun@itp.ac.cn}{\texttt{chenkun@itp.ac.cn}}.}}
\affil[1]{Department of Modern Physics, University of Science and Technology of China, Hefei, Anhui 230026, China}
\affil[2]{CAS Key Laboratory of Theoretical Physics, Institute of Theoretical Physics, Chinese Academy of Sciences, Beijing 100190, China}
\affil[3]{Hefei National Laboratory, University of Science and Technology of China, Hefei 230088, China}
\affil[4]{Hefei National Research Center for Physical Sciences at the Microscale and School of Physical Sciences, University of Science and Technology of China, Hefei 230026, China}
\affil[5]{Institute for Advanced Algorithms Research, Shanghai 200120, China}
\affil[6]{Endless Frontier, Shanghai 200030, China}
\date{}

\begin{document}
\maketitle
\begin{abstract}
The next frontier for artificial general intelligence is tackling unresolved scientific problems, calling for benchmarks that assess progress beyond established knowledge. We introduce OpenProblemBench, a benchmark of 82 unresolved problems drawn from the mathematics and theoretical physics literature. Each problem supplies the research context, assumptions, and prior progress needed to investigate the question. We select problems whose proposed solutions admit comparatively clear checks of their decisive mathematical or computational claims. Four evaluator models independently assess the correctness, completeness, and degree of progress of each submission without reference solutions. Across seven evaluated configurations, GPT-6-Astra achieves the highest mean judged solve rate of 14.0\%, compared with 5.5--6.7\% for the evaluated full-size open models and 2.4--3.7\% for Flash models. Case comparisons connect stronger outcomes to changes in problem representation, general arguments that extend beyond finite evidence, and proofs of the steps needed to complete a solution. By grounding evaluation in questions arising from the research literature, OpenProblemBench provides a setting for investigating the capabilities and limitations of AI as a contributor to foundational theoretical science.
\end{abstract}

\section{Introduction}
The strongest contemporary language models can sustain long investigations, combine symbolic reasoning with computation, revise failed approaches, and develop mathematical proofs and scientific arguments. These capabilities are beginning to matter beyond routine research assistance. Recent reports describe an AI-generated finite-time blowup proof for the three-dimensional incompressible Navier--Stokes equations with smooth external forcing, accompanied by a Lean formalization \citep{openai2026ns}, and an AI-generated counterexample to a conjecture in discrete geometry \citep{openai2026geometry}. Such results raise a question that conventional knowledge tests cannot answer: can AI advance research questions that remain unresolved in the expert literature?

Most scientific benchmarks stop short of this boundary. They test advanced reasoning, coding, data analysis, or the reconstruction of published findings, all of which are necessary components of research \citep{tian2024scicode,chen2025scienceagentbench,wang2026frontierscience,starace2025paperbench}. These tasks usually have a published result or a known target against which progress can be measured. For an unresolved problem, the final answer is unavailable. The system must decide which direction is productive, determine what evidence would close the argument, and recognize when a plausible derivation still leaves a quantifier, parameter regime, or exceptional case untreated.

Open problems are therefore a stringent test of research capability: they lie beyond the frontier documented by the relevant expert literature at the time of curation. At the same time, useful progress can often be assessed more objectively than the open-ended nature of the task suggests. A proposed counterexample can be checked against the stated assumptions; a classification can be tested for necessity and sufficiency; a computational certificate can be reproduced; and a proof can be inspected at its decisive steps. This combination---high solution difficulty with comparatively structured verification---makes selected open problems suitable for evaluating research-oriented systems.

We introduce \bench{}, a benchmark of 82 literature-grounded problems in mathematics and theoretical physics. Each item states the unresolved objective, relevant context, assumptions, known progress, and the conditions that a resolution must satisfy. Four evaluator models independently review each proposed solution and distinguish solved outcomes from trivial, nontrivial, and breakthrough partial progress. The resulting solver--evaluator matrix measures both complete solutions and useful advances on questions for which no reference answer is available.

The initial study evaluates seven configurations on all 82 problems, producing 574 final submissions and 2,296 reviews. GPT-6-Astra leads every evaluator and attains a mean judged solve rate of 14.02\%, while all other configurations remain below 8\%. The wider outcome distribution is equally informative: Qwen3.8-Max and Kimi-K3 often receive substantive-progress judgments without resolving the full problem, and GLM-5.3-Flash receives solved judgments while approaching GLM-5.3 in the combined rate of solved, breakthrough, and nontrivial outcomes. Selected case comparisons show how changing a problem's representation can support a general proof. They also identify why other attempts remain partial: finite evidence, an unproved necessary lemma, or checks that share the same mathematical error.

\bench{} contributes a benchmark at the boundary of established knowledge, a protocol for assessing completion and partial progress without reference solutions, and a comparison of how different systems approach the same research questions. The comparison examines whether systems can select productive directions, develop new arguments, and establish progress on questions beyond the current expert literature.

\section{Related Work}
\label{sec:related}
\subsection{Scientific reasoning with known targets}
FrontierMath evaluates advanced mathematical reasoning through automatically checked answers, while FrontierScience includes Olympiad and doctoral-level tasks across physics, chemistry, and biology \citep{glazer2024frontiermath,wang2026frontierscience}. SciCode focuses on research coding, and ScienceAgentBench reconstructs data-driven workflows from scientific publications \citep{tian2024scicode,chen2025scienceagentbench}. These benchmarks measure demanding prerequisites for research, but the expected result is available when the task and its evaluation are constructed.

Several newer suites broaden the unit of evaluation from an answer to a research process. AstaBench studies scientific tasks under different tool settings, SciAgentArena evaluates interactive challenges across scientific scales, and AutoResearchEval analyzes complete trajectories \citep{bragg2026astabench,liu2026arena,fei2026autoresearch}. FIRE-Bench asks systems to rediscover established insights, while TruthInsightBench evaluates how well claims are supported by evidence on study-derived tasks \citep{wang2026fire,yang2026truth}. These designs reveal how a system searches, computes, and reports evidence; \bench{} retains these concerns but places them against targets whose final answers are not known.

\subsection{Research reproduction and generation}
The AI Scientist demonstrates an end-to-end system that proposes machine-learning ideas, runs experiments, writes papers, and invokes automated review \citep{lu2024aiscientist}. PaperBench tests the replication of 20 recent machine-learning papers with author-developed rubrics, separately evaluates its automated judge, and includes a human baseline; PRL-Bench reconstructs exploratory theoretical and computational tasks from 100 recent \emph{Physical Review Letters} papers \citep{starace2025paperbench,miao2026prl}. Terminal-Bench-Science spans 70 research workflows, whereas MLRC-Bench evaluates methodological advances through empirical performance \citep{dillmann2026tbscience,zhang2025mlrc}. These benchmarks provide rich, research-scale tasks with a published result or empirical objective. OpenProblemBench evaluates progress on questions left unresolved by the source literature.

\subsection{Evaluation on unresolved questions}
HorizonMath contains 113 predominantly unsolved problems across mathematical and mathematical-science domains. Its answers are computational objects or expressions evaluated through admissibility, construction, and numerical checks \citep{wang2026horizon}. FrontierMath: Open Problems similarly emphasizes verifiable mathematical challenges \citep{epoch2026open}. UQ draws 500 unanswered questions from online communities and combines automated screening with community verification \citep{nie2025uq}. OpenProblemBench extends open-problem evaluation through a systematic collection of research questions in theoretical mathematics and physics, each connected to the assumptions and unresolved conclusions of prior work. We select problems with comparatively clear procedures for checking proposed solutions and use multiple agent judges to review the arguments independently. This approach supports evaluation without requiring formalization in Lean or another proof assistant. Table~\ref{tab:related} compares the evaluation designs.

\section{Benchmark Construction and Evaluation}
\label{sec:method}
\subsection{Literature-grounded problem collection}
We identify candidate open research questions from literature-derived information provided by the Large Knowledge Model (LKM) \citep{huang2026lkm}. We then review the source papers and subsequent literature to assess whether the questions remain open and organize the retained questions into self-contained problem statements.

This process yields 82 open research problems in mathematics and theoretical physics. Each retained statement is checked for fidelity to the source literature and scientific relevance. The literature review did not identify a complete solution at the time of curation.

Each solver receives the problem background, problem statement, scientific significance, current progress, and references. We favor questions whose proposed solutions admit comparatively objective scrutiny through inspection of decisive proof steps, validation of constructions against their assumptions, or reproduction of consequential calculations.

\subsection{Composition and scope}
The benchmark contains 46 mathematics and 36 physics problems. Mathematics covers arithmetic questions, algebraic geometry, topology, analysis, dynamics, combinatorics, optimization, and logic. Physics covers statistical mechanics and integrable systems, quantum information, gauge and field theory, optics, and algebraic or combinatorial mathematical physics. Table~\ref{tab:coverage} gives a primary-subfield grouping; overlapping subjects are assigned once for counting. The original domain labels are retained, including for physics problems with substantial mathematical content.

The questions range from recent conjectures to problems that have remained open for centuries. The Goldbach item traces its origin to 1742, whereas the iterated-fusion item comes from work first posted in 2025 and published in 2026 \citep{ikhlef2026fusion}. The benchmark therefore spans gaps of roughly one year to more than 280 years.

\subsection{Solver configurations and execution}
\label{sec:execution}
The primary comparison uses seven models and one final submission per model--problem pair, for 574 submissions. GPT-6-Astra and GPT-5.6-Sol run in Codex CLI; DeepSeek-V4.1-Flash runs in DeepSeek Harness; Kimi-K3, GLM-5.3, Qwen3.8-Max, and GLM-5.3-Flash run in OpenCode. Figures use model names, and the text abbreviates these as Astra, GPT-Sol, Kimi, GLM, Qwen, GLM-Flash, and DeepSeek-Flash.

Each run produces a self-contained solution that is expected to distinguish established results, conditional conclusions, heuristics, and remaining gaps. Local computation is available and network retrieval is disabled in the primary setting. The documented protocol permits up to 12 hours of sustained work without a fixed token budget; realized use varies. The evaluation applies to the resulting model--execution configurations rather than model weights in isolation.

An additional Qwen condition supports a paired comparison of OpenCode and Claude Code, with network access disabled in both. The comparison uses the same 82 statements and GLM-5.3 evaluator.

\subsection{Agent-as-judge assessment}
Four evaluator models---GPT-5.6-Sol, Kimi-K3, GLM-5.3, and Qwen3.8-Max---review each frozen submission independently against the exact statement shown to the solver. An evaluator first identifies the obligations imposed by the problem, then checks assumptions, quantifiers, decisive proof steps, and supporting computations. Targeted recomputation and source checks are permitted under the evaluator's access policy. The evaluator must assess the submitted argument as written, without filling a missing lemma or importing another evaluator's conclusion, and must record both established contributions and remaining gaps.

A completed review assigns one of three outcomes. \emph{Solved} means that all task requirements are established. \emph{Partial} means that at least one relevant result beyond the supplied premises is established, while some requirements remain open. \emph{Unsolved} means that no such result is established. Partial results receive one of three qualitative levels: \emph{trivial}, a direct standard application or easy special case; \emph{nontrivial}, a meaningful family, improved bound, or useful reduction that removes part of the difficulty; or \emph{breakthrough}, a result that resolves a central obstacle and substantially changes the remaining work. These levels describe progress relative to the task. A review is marked \emph{Incomplete} when the evaluator reports insufficient evidence to make a resolution judgment.

Review reports also record overclaiming when a submission asserts more completion, generality, or verification than its argument supports. We use these findings as a diagnostic of how reliably a model describes its own open-problem work.

Each verdict cites the relevant task requirements and the mathematical or computational evidence supporting its assessment. These reports allow readers to trace disagreements to specific claims and reproduce the associated checks.

\subsection{Reported outcomes}
For each solver--evaluator pair, the solve rate is the number of problems judged solved divided by 82. We also report the three partial-progress levels, nonsolved outcomes, and Incomplete assessments. We use \emph{substantive partial progress} to refer jointly to nontrivial and breakthrough outcomes. All 28 primary solver--evaluator cells contain 82 reviews.

For the overall comparison, each evaluator receives equal weight and we average the percentage in each outcome category across the four evaluators. Figure~\ref{fig:mean} presents solved, breakthrough partial, nontrivial partial, trivial partial, and nonsolved or incomplete outcomes. Figure~\ref{fig:partial-overclaim} focuses on the share of nontrivial and breakthrough outcomes among completed reviews not judged solved, and reports detected overclaims among reviews for which that diagnostic is determined. Appendix~\ref{app:counts} gives the underlying counts.

\section{Results and Analysis}
\label{sec:results}
\subsection{Overall performance across evaluators}
\begin{figure}[t]
\centering\includegraphics[width=\linewidth]{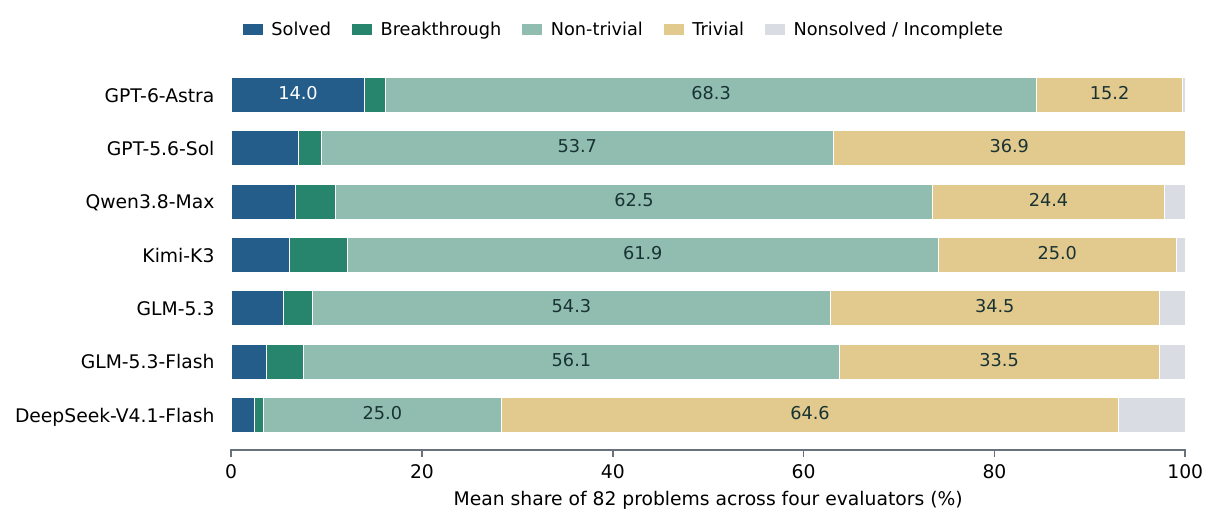}
\caption{Outcome distribution averaged equally across the four evaluators and ordered by mean judge-assessed solve rate. Each evaluator uses all 82 problems as the denominator. The final segment combines completed nonsolved verdicts with incomplete assessments, for which the evaluator reported insufficient evidence to make a resolution judgment. Raw counts retain these two outcomes separately.}
\label{fig:mean}
\end{figure}

Figure~\ref{fig:mean} gives the primary comparison by averaging each outcome category equally across the four evaluators. Astra has the highest mean judged solve rate, 14.02\%, and the largest share of outcomes rated solved, breakthrough, or nontrivial, 84.45\%. Solved judgments remain rare for every other configuration: their mean rates range from 2.44\% to 7.01\%. The distribution of partial outcomes reveals differences hidden by solve rate. GPT-Sol receives more solved judgments than Qwen or Kimi, while Qwen and Kimi receive solved or substantive partial judgments in 73.48\% and 74.09\% of reviews, compared with 63.11\% for GPT-Sol. Kimi also has the largest breakthrough-partial segment, 6.10\%, indicating that several of its strongest attempts are assessed as removing a central obstacle without closing the full problem.

GLM-Flash receives fewer solved judgments than GLM on average, 3.66\% versus 5.49\%, yet their combined solved, breakthrough, and nontrivial shares are close, 63.72\% and 62.80\%. GLM-Flash thus produces substantial intermediate results at a similar rate to GLM, with a lower rate of complete solutions. DeepSeek-Flash is dominated by trivial partial outcomes (64.63\%), although it too receives solved judgments. The distribution distinguishes models that complete more problems from those that make useful intermediate progress.

\begin{figure}[t]
\centering\includegraphics[width=.95\linewidth]{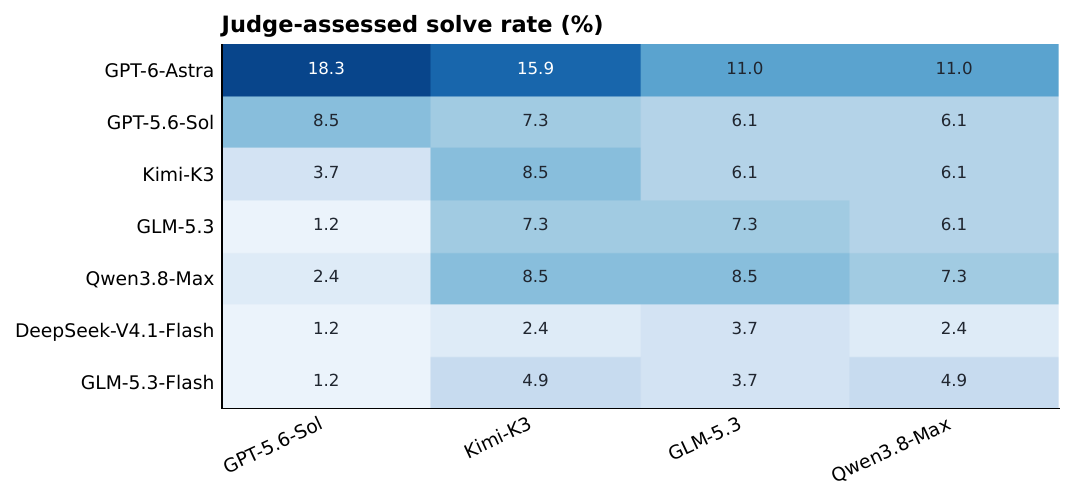}
\caption{Solve rates under each evaluator for the seven primary configurations. Each cell uses 82 problems; incomplete assessments contribute no successes. Execution environments are specified in Section~\ref{sec:execution}.}
\label{fig:solve}
\end{figure}

The evaluator-specific matrix in Figure~\ref{fig:solve} shows that Astra leads in every column, with 15, 13, 9, and 9 solved submissions. Rankings below the leader are less stable. Under Qwen, for example, GLM-Flash receives four solved judgments, GPT-Sol, Kimi, and GLM receive five each, and Qwen receives six. The union across configurations contains between 13 and 16 problems judged solved by a fixed evaluator. These subsets overlap only partially, making per-problem comparisons more informative than a single aggregate ordering.

\subsection{Partial progress and reporting reliability}
\begin{figure}[t]
\centering\includegraphics[width=.95\linewidth]{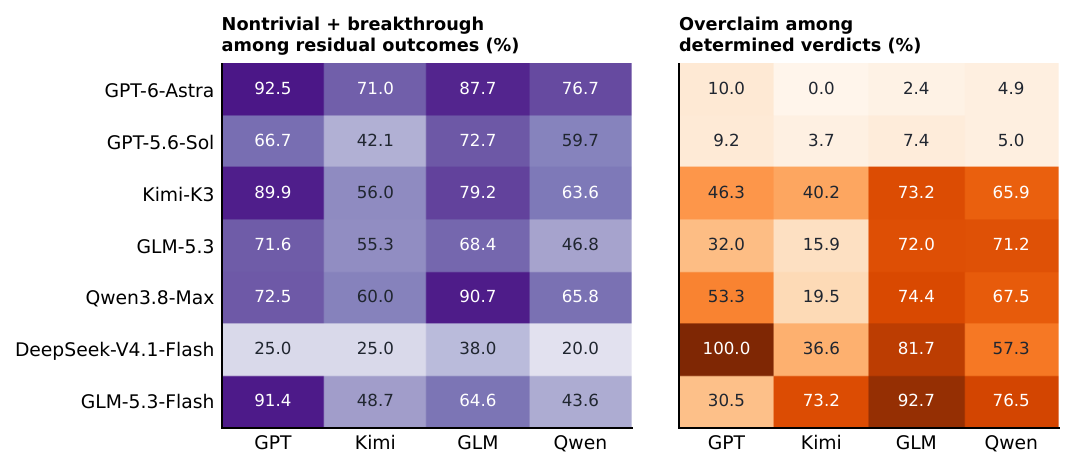}
\caption{Partial progress and overclaim diagnostics under each evaluator. Left: nontrivial plus breakthrough outcomes among completed reviews not judged solved. Right: detected overclaims among reviews with a determined overclaim verdict; diagnostic coverage is reported in Appendix~\ref{app:counts}.}
\label{fig:partial-overclaim}
\end{figure}

Of 2,296 primary reviews, 2,288 contain a completed outcome assessment. The 2,095 partial judgments comprise 768 trivial, 1,252 nontrivial, and 75 breakthrough outcomes. Because these are review-level counts, evaluators can assign different progress levels to the same submission. The left panel of Figure~\ref{fig:partial-overclaim} asks, among completed reviews not judged solved, what fraction are rated nontrivial or breakthrough. Astra remains high under all evaluators (71.0--92.5\%), while Qwen and Kimi vary more strongly across evaluators (60.0--90.7\% and 56.0--89.9\%); DeepSeek-Flash is lower (20.0--38.0\%). These ranges show how the distinction between trivial and substantive progress varies across evaluators, alongside differences between solvers.

The right panel summarizes whether models accurately state what their arguments establish. Typical findings include treating a conditional derivation as unconditional, extrapolating a finite check to an infinite family, or omitting an exceptional parameter stratum from a claimed classification. These findings describe how reliably models report the scope and completeness of their results. Because the diagnostic is not determined for every review, Appendix~\ref{app:counts} lists its coverage for each solver--evaluator pair.

\subsection{Harness and web-access sensitivity}
Table~\ref{tab:ablations} reports two comparisons of Qwen3.8-Max on the same 82 statements: OpenCode versus Claude Code with solver network access disabled, and offline versus web-enabled solving within Claude Code. GLM-5.3 assesses all submissions using the same offline review protocol, so the web comparison changes the solver's access to external information while keeping reviewer access fixed.

\begin{table}[t]
\centering\small
\caption{Qwen3.8-Max harness and web-access comparisons on the same 82 statements, with GLM-5.3 using the same offline review protocol. Web access refers to the solver. Nontrivial + breakthrough counts partial results only; all percentages use 82 as the denominator.}
\label{tab:ablations}
\begin{tabular}{@{}llrr@{}}\toprule
Harness & Web access & Solved $n$ (\%) & Nontrivial + breakthrough $n$ (\%) \\\midrule
OpenCode & No & 7 (8.54) & 68 (82.93) \\
Claude Code & No & 5 (6.10) & 59 (71.95) \\
Claude Code & Yes & 5 (6.10) & 68 (82.93) \\
\bottomrule
\end{tabular}
\end{table}

Changing from OpenCode to Claude Code reduces the solved count from 7 to 5 and substantive partial outcomes from 68 to 59. One problem gains a solved judgment and three lose it; Appendix~\ref{app:ablations} identifies these four problems. Within Claude Code, enabling web access leaves the solved count unchanged at 5 but raises substantive partial outcomes from 59 to 68, including an increase in breakthrough outcomes from 4 to 7. The solved sets differ: three problems gain a solved judgment and three lose it.

These results show that execution settings affect performance on individual problems and the extent of partial progress. In this Qwen comparison, web access supports more substantive progress on open problems without increasing the overall judged solve rate. Astra retains its lead under the same evaluator across the tested Qwen settings.

\subsection{How models approach the same open problems}
\label{sec:approaches}
The aggregate results leave a practical question: what separates a complete argument from substantial but unfinished work on the same problem? Comparing six models on ten shared questions reveals four recurring differences in their submitted solutions.

\paragraph{Reformulating the problem.} ORB-MATH-30 asks for a parameterized volume-conjecture asymptotic for knot polynomials throughout a complex neighborhood. Astra follows the permitted refutation route: bounds at roots of unity and an explicit sequence of irrational parameters contradict the requested asymptotic. Qwen and Kimi develop computational pipelines for colored knot polynomials and examine numerical and branch ambiguities, but stop without a proof or counterexample. On ORB-PHYS-57, which asks for the exact critical surface of a bow-tie percolation lattice, Astra replaces an invalid local probability interpretation with an identity on finite tori; Qwen's comparison bounds and simulations leave a gap to the critical surface. In these cases, Astra finds a representation that directly addresses the requested conclusion, while the other attempts improve control over intermediate quantities.

\paragraph{From finite evidence to general proof.} ORB-PHYS-62 asks whether the error in Pauling's residual-entropy estimate vanishes as graph degree grows, uniformly over graph size. Qwen proves a regime that still relates size to degree, Kimi reduces a structured family to an unproved local-limit estimate, and GLM-Flash derives exact fixed-degree transfer operators. Astra supplies a bound independent of graph size and is judged solved by all four evaluators. ORB-PHYS-69, the classification of Sklyanin algebras over arbitrary fields, exposes a similar gap. Qwen, GLM-Flash, and DeepSeek-Flash use exact finite-field searches to find algebras that become isomorphic after a field extension. Their general classification remains incomplete because they do not exclude all further isomorphisms. Exact computation establishes useful cases, but a general argument is still needed to cover every admissible instance.

\paragraph{Closing gaps in the proposed proof.} ORB-PHYS-71 asks whether iterated binary fusion is isomorphic to a directly defined $m$-fold fusion. Astra proves that the direct construction is well defined, then uses a common planar convolution representation to construct mutually inverse maps; all four evaluators judge its solution solved. Other submissions leave essential steps unproved: Kimi leaves normalization and relation transfer unresolved, GLM leaves injectivity conditional, and GLM-Flash introduces an underived rule. These gaps prevent the submitted arguments from establishing the requested isomorphism. On ORB-PHYS-76, concerning flow and tension counting on non-orientable surfaces, Kimi's tension implementation and brute-force comparator share an incorrect dual convention; an independent twisted-bridge identity exposes the error. GLM checks duality and known planar counts, while GLM-Flash corrects its representation after detecting errors; both receive solved judgments from Qwen. The contrast is whether a submission establishes the required mathematical correspondence, beyond agreement between implementations.

\paragraph{Constructive reasoning in Flash models.} Comparing GLM with GLM-Flash shows how a Flash model can also develop effective research arguments. In the entropy and percolation cases above, GLM builds explicit mathematical models but leaves a uniform estimate or boundary extension unresolved. GLM-Flash constructs exact witnesses for the algebra-classification problem and combines valuation arguments with symbolic elimination on ORB-PHYS-82, which asks when generalized lambda-function values are integral. Its solution to the latter problem is judged solved by Qwen, as is GLM's. Its fusion attempt also shows where construction falls short: proving an isomorphism in a modified framework leaves the original task unresolved. Together with its complete counting-recurrence solution, these cases illustrate the potential of Flash models to advance open questions, while identifying gaps in their ability to derive auxiliary assumptions, cover exceptional cases, and complete a proof.

\subsection{Disciplinary differences and evaluator agreement}
Under Qwen evaluation, Astra solves one of 46 mathematics problems and eight of 36 physics problems; no other model receives a solved mathematics judgment. This gap reflects the sampled tasks as much as disciplinary capability. Requiring comparatively checkable resolutions selects broad, longstanding conjectures on the mathematics side, whereas many physics items ask for extensions of a specified construction. Several physics items, including fusion and modular-function integrality, also demand substantial mathematics. The present comparison therefore reflects this task composition rather than general disciplinary proficiency. Extending coverage to physical modeling, empirical interpretation, and other sciences would test abilities that these tasks do not capture.

Ten problems have a submission judged solved by all four evaluators, eight from Astra; eighteen have at least one solved judgment from at least one evaluator. Appendix~\ref{app:agreement} identifies these shared and evaluator-dependent outcomes for follow-up verification.

\section{Conclusion}
\bench{} evaluates AI on unresolved research questions in mathematics and theoretical physics. Across 82 problems, Astra has the highest solve rate under every evaluator, while several other models make substantive partial progress and Flash models solve a subset of the tasks. The case comparisons connect these outcomes to concrete research decisions: changing the representation of a problem, extending finite results to a general argument, and checking the step that establishes the full conclusion. By testing progress where established answers are unavailable, the benchmark assesses the capacity of AI systems to explore beyond existing human knowledge and develop justified scientific conclusions.

\section{Limitations and Future Work}
\label{sec:limits}
The results are model judgments; independent expert review is needed to establish scientific correctness and novelty. Evaluators may share errors, and their agreement alone does not estimate false positive or false negative rates. A validation release should use blinded domain experts on a stratified sample spanning unanimous solved judgments, disagreements, partial-progress levels, overclaim findings, and all-negative cases. Formal proofs, executable constructions, reproduced calculations, and a fresh literature review can further support candidate discoveries.

The 82 questions are uneven across subfields: the checkability criterion selects broad mathematics problems and a mathematically intensive subset of theoretical physics. Literature searches may miss solutions, and relevant statements may occur in training data. Future releases should specify a literature cutoff, add human-researcher baselines, and extend coverage to chemistry, biology, and economics.

Historical runs also vary in resources and execution software, and a single run per ablation condition cannot measure run-to-run variance. Fine comparisons require repeated runs under common budgets.


\bibliography{references}
\bibliographystyle{plainnat}
\clearpage
\appendix
\FloatBarrier
\section{Comparison of evaluation designs}
\begin{table}[h!]
\centering\small
\caption{Representative evaluation designs, distinguishing tasks with published targets from unresolved research questions.}
\label{tab:related}
\begin{tabularx}{\linewidth}{@{}>{\raggedright\arraybackslash}p{.21\linewidth}YY@{}}\toprule
Benchmark & Evaluation target & Evidence / assessment \\\midrule
SciCode; ScienceAgentBench & Research coding and publication-derived workflows & Reference implementations, tests, and execution outputs \\
FrontierScience & Expert scientific tasks and research subtasks & Scientist-authored answers and granular criteria \\
AstaBench; AutoResearchEval; TB-Science & Scientific workflows and agent behavior & Task-specific outcomes, tools, and process analysis \\
FIRE-Bench; TruthInsightBench & Insight rediscovery / evidence-grounded discovery on study-derived tasks & Empirical findings / claim-level evidence maturity \\
PaperBench; PRL-Bench & Replication of machine-learning papers / exploratory physics tasks & Author-developed rubrics / reference answers and expert rubrics \\
MLRC-Bench & Open-ended machine-learning methodology & Empirical research performance \\
HorizonMath & Predominantly unsolved mathematical discovery & Computational objects and numerical/construction checks \\
FrontierMath: Open Problems & Unsolved mathematical challenges & Problem-specific verifiers \\
UQ & Unanswered community questions across domains & Validator screening and community verification \\
\textbf{OpenProblemBench} & \textbf{Unresolved theoretical questions from prior research} & \textbf{Argument-based assessment by multiple independent judges} \\\bottomrule
\end{tabularx}
\end{table}

\FloatBarrier
\section{Detailed Outcome and Diagnostic Counts}
Of the 2,296 primary reviews, eight are marked Incomplete: five GPT-Sol reviews of DeepSeek-Flash and three Qwen reviews of Qwen. All other primary reviews have a completed outcome assessment.
\label{app:counts}
\begin{table}[h]
\centering\small
\caption{Submissions judged solved out of 82, with percentages in parentheses. All cells contain 82 review records; incomplete outcome assessments are reported separately.}
\begin{tabular}{@{}lrrrr@{}}\toprule
Model & GPT & Kimi & GLM & Qwen \\\midrule
GPT-6-Astra & 15 (18.29\%) & 13 (15.85\%) & 9 (10.98\%) & 9 (10.98\%) \\
GPT-5.6-Sol & 7 (8.54\%) & 6 (7.32\%) & 5 (6.10\%) & 5 (6.10\%) \\
Kimi-K3 & 3 (3.66\%) & 7 (8.54\%) & 5 (6.10\%) & 5 (6.10\%) \\
GLM-5.3 & 1 (1.22\%) & 6 (7.32\%) & 6 (7.32\%) & 5 (6.10\%) \\
Qwen3.8-Max & 2 (2.44\%) & 7 (8.54\%) & 7 (8.54\%) & 6 (7.32\%) \\
DeepSeek-V4.1-Flash & 1 (1.22\%) & 2 (2.44\%) & 3 (3.66\%) & 2 (2.44\%) \\
GLM-5.3-Flash & 1 (1.22\%) & 4 (4.88\%) & 3 (3.66\%) & 4 (4.88\%) \\
\bottomrule\end{tabular}
\end{table}
\begin{table}[h]
\centering\small
\caption{Nontrivial / breakthrough / all partial counts. Categories include completed assessments only; these are review events, not unique discoveries.}
\begin{tabular}{@{}lrrrr@{}}\toprule
Model & GPT & Kimi & GLM & Qwen \\\midrule
GPT-6-Astra & 60/2/67 & 48/1/69 & 62/2/73 & 54/2/72 \\
GPT-5.6-Sol & 49/1/75 & 30/2/76 & 54/2/77 & 43/3/77 \\
Kimi-K3 & 62/9/78 & 39/3/74 & 57/4/77 & 45/4/76 \\
GLM-5.3 & 57/1/80 & 38/4/74 & 49/3/76 & 34/2/71 \\
Qwen3.8-Max & 56/2/79 & 41/4/75 & 62/6/75 & 46/2/70 \\
DeepSeek-V4.1-Flash & 17/2/65 & 19/1/79 & 30/0/79 & 16/0/74 \\
GLM-5.3-Flash & 67/7/81 & 35/3/77 & 49/2/79 & 33/1/70 \\
\bottomrule\end{tabular}
\end{table}
\begin{table}[h]
\centering\small
\caption{Overclaim counts: detected / not detected / unknown. Unknown means that the archived diagnostic did not establish a complete yes/no verdict, usually because coverage was partial. Labels are recovered from linked diagnostic files or explicit archived report verdicts. A dash means no reviews for the entire cell.}
\begin{tabular}{@{}lrrrr@{}}\toprule
Model & GPT & Kimi & GLM & Qwen \\\midrule
GPT-6-Astra & 7/63/12 & 0/82/0 & 2/80/0 & 4/78/0 \\
GPT-5.6-Sol & 7/69/6 & 3/79/0 & 6/75/1 & 4/76/2 \\
Kimi-K3 & 38/44/0 & 33/49/0 & 60/22/0 & 54/28/0 \\
GLM-5.3 & 8/17/57 & 13/69/0 & 59/23/0 & 57/23/2 \\
Qwen3.8-Max & 16/14/52 & 16/66/0 & 61/21/0 & 52/25/5 \\
DeepSeek-V4.1-Flash & 18/0/64 & 30/52/0 & 67/15/0 & 47/35/0 \\
GLM-5.3-Flash & 25/57/0 & 60/22/0 & 76/6/0 & 62/19/1 \\
\bottomrule\end{tabular}
\end{table}
\begin{table}[h]
\centering\small
\caption{Overclaim diagnostic coverage: determined verdicts as a percentage of 82 submissions. The complement is the unknown share.}
\begin{tabular}{@{}lrrrr@{}}\toprule
Model & GPT & Kimi & GLM & Qwen \\\midrule
GPT-6-Astra & 85.4\% & 100.0\% & 100.0\% & 100.0\% \\
GPT-5.6-Sol & 92.7\% & 100.0\% & 98.8\% & 97.6\% \\
Kimi-K3 & 100.0\% & 100.0\% & 100.0\% & 100.0\% \\
GLM-5.3 & 30.5\% & 100.0\% & 100.0\% & 97.6\% \\
Qwen3.8-Max & 36.6\% & 100.0\% & 100.0\% & 93.9\% \\
DeepSeek-V4.1-Flash & 22.0\% & 100.0\% & 100.0\% & 100.0\% \\
GLM-5.3-Flash & 100.0\% & 100.0\% & 100.0\% & 98.8\% \\
\bottomrule\end{tabular}
\end{table}

\FloatBarrier
\section{Cross-judge Solved Outcomes}
\label{app:agreement}
\begin{figure}[h]
\centering\includegraphics[width=.95\linewidth]{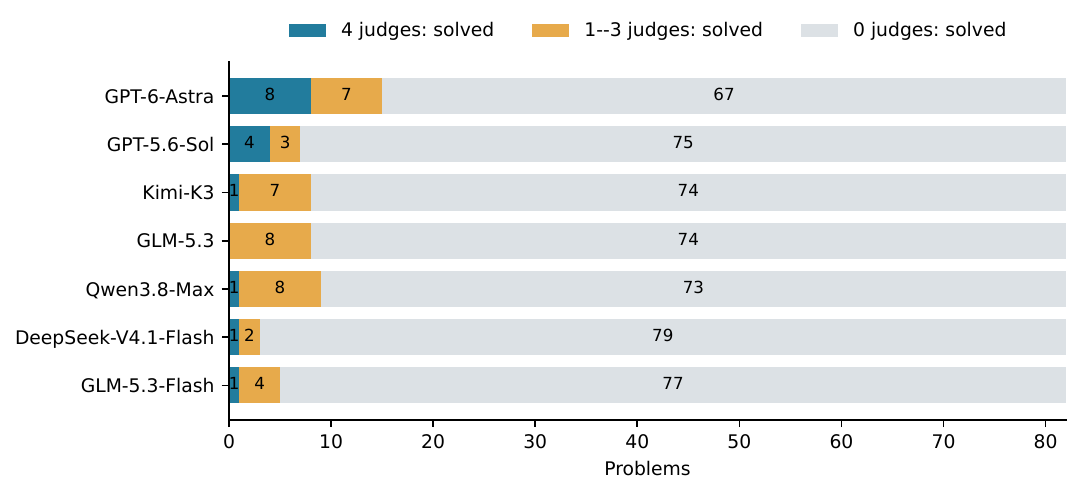}
\caption{Numbers of problems judged solved by zero, one to three, or four evaluators for each configuration. The zero category includes submissions with Incomplete assessments and no solved judgment.}
\end{figure}
For Astra, eight submissions are judged solved unanimously, seven receive one to three solved judgments, and 67 receive none. The unanimous set comprises ORB-MATH-30 and ORB-PHYS-56, ORB-PHYS-57, ORB-PHYS-62, ORB-PHYS-64, ORB-PHYS-69, ORB-PHYS-71, and ORB-PHYS-76. The cross-configuration unanimous-solved union additionally contains ORB-PHYS-77 and ORB-PHYS-82. The corresponding arguments and reviewer reports provide a focused set for independent scientific adjudication.

\FloatBarrier
\section{Paired Harness Comparison}
\label{app:ablations}
The harness comparison uses Qwen3.8-Max submissions to the same 82 statements, with network access disabled and GLM-5.3 assessing each submission. ORB-PHYS-56 is judged breakthrough under OpenCode and solved under Claude Code. ORB-PHYS-50, ORB-PHYS-64, and ORB-PHYS-77 are judged solved under OpenCode and breakthrough under Claude Code.

We use the exact conditional form of McNemar's paired test \citep{mcnemar1947}. Given $g$ gained and $l$ lost solve outcomes when moving from OpenCode to Claude Code, the null hypothesis gives equal probability to either direction among the $g+l$ discordant pairs. Here $(g,l)=(1,3)$: conditional on four changes, the probability of a split at least this uneven is $2[\binom{4}{0}+\binom{4}{1}]/2^4=0.625$. This test measures the directional balance of changes in solve status. At the broader threshold of solved or at least nontrivial progress, OpenCode and Claude Code have 75 and 64 outcomes, respectively.

We verified that all 82 statement hashes match across the two conditions and that all 164 reviews are complete and linked to the corresponding submissions. The comparison characterizes the archived execution configurations; repeated runs under common resource budgets would be needed to isolate harness effects from run-to-run variation.

\FloatBarrier
\section{Detailed Shared-problem Comparisons}
\label{app:cases}
\subsection{Selection and comparison procedure}
Section~\ref{sec:approaches} compares Astra, Qwen, Kimi, GLM, GLM-Flash, and DeepSeek-Flash on ten questions selected for contrasting outcomes. We read their final solutions, all four evaluator reports for each submission, and supporting calculations relevant to the claimed results. This qualitative sample supports comparisons of observed approaches, not estimates of their frequency across the benchmark. This appendix develops four examples in detail; its table holds the evaluator fixed to Qwen and additionally includes GPT-Sol. These examples illustrate the proof obligations, intermediate contributions, and remaining gaps used to interpret the contrasting evaluator judgments.

\begin{table}[htbp]
\centering\small
\caption{Contrasting approaches on four research problems, assessed by the same Qwen reviewer. S: solved; B: breakthrough partial; N: nontrivial partial; T: trivial partial. Model abbreviations refer to the primary configurations in Section~3.3. Problem content and proof obligations are described below.}
\label{tab:cases}
\setlength{\tabcolsep}{4pt}
\begin{tabular}{@{}lccccccc@{}}\toprule
Problem ID & Astra & GPT-5.6 & Kimi & GLM & Qwen & \shortstack{GLM\\Flash} & \shortstack{DeepSeek\\Flash} \\\midrule
ORB-PHYS-57 & S & N & S & B & N & N & T \\
ORB-PHYS-62 & S & N & N & N & N & N & T \\
ORB-PHYS-69 & S & S & B & N & B & N & N \\
ORB-PHYS-82 & S & S & S & S & S & S & S \\
\bottomrule
\end{tabular}
\end{table}

\subsection{ORB-PHYS-57: Exact critical manifold of the inhomogeneous bow-tie lattice}
The problem asks whether the five-parameter critical surface proposed by \citet{ziff2012bowtie} locates the percolation transition throughout the admissible domain, including the regime in which a split-bond parameter is negative. Recovering this polynomial algebraically is insufficient: a negative parameter does not define an ordinary probability law, and the connection to the infinite-volume phase transition must be justified.

Astra's Sections 2--4 document a change in the object being controlled. A single signed triangle cannot be interpreted as a positive local random generator, and direct stochastic comparison leaves a gap. Instead, the solution groups transformations on finite tori: signed intermediate edges cancel after degree-two vertices are suppressed, leaving the complementary dual with genuine probabilities. An explicit decorated-graph isomorphism gives the finite-volume winding identity; separately stated sharpness and noncoexistence results then locate the transition. Kimi's Sections 3--6 take a related finite-volume route through wrap polynomials, star--triangle equivalence, and recombination. Both are judged solved by Qwen.

Qwen's own attempt pursues comparison bounds, duality equations, and counterexample searches. It builds weighted non-backtracking matrices for the lattice and its dual to bracket the critical parameter; Monte Carlo is explicitly supporting evidence. This provides useful control but does not collapse the upper and lower bounds to the claimed surface. DeepSeek-Flash similarly retains algebraic identities and one-sided bounds and is rated trivial partial. GLM gets further: its work record documents replacing an invalid crossing-event identification with a decay-transfer argument, but Qwen retains breakthrough status because full-domain completion is missing. The contrast identifies a useful research behavior: change from an insufficient local probability interpretation to an exact finite-system identity with a valid limiting argument.

\subsection{ORB-PHYS-62: Pauling's residual entropy estimate at growing degree}
For a graph $G$ with $n$ vertices, residual entropy is $\rho(G)=n^{-1}\log \mathrm{EO}(G)$, where $\mathrm{EO}(G)$ counts orientations with equal in-degree and out-degree at every vertex. For simple $d$-regular graphs with even $d$, the task asks whether the error in Pauling's estimate $\widehat\rho_d=\log\binom{d}{d/2}-(d/2)\log 2$ vanishes as $d\to\infty$, uniformly over graph size and without the structural assumptions in prior results \citep{hasheminezhad2025pauling}. A family of matching numerical examples, or a result imposing an additional relationship between vertex count $n$ and degree $d$, leaves part of this target open.

Astra's Section 3 records successive approaches: circuit-partition identities, entropy bounds, Fourier estimates for large complete blocks, and finally spectral truncation. Its record explicitly explains why the earlier bounds do not vanish or cover only special cases. Section 10 separates a low-rank exceptional spectral subspace and controls the remaining Fourier integral, giving an $n$-independent bound $O((\log d/d)^{1/3})$. The accompanying audit checks normalizations, scalar inequalities, and values of the explicit bound; it supports the submitted analytic argument rather than extrapolating a theorem from graph samples. All four evaluators judge this answer solved.

Kimi constructs cycle blow-ups that evade the known sufficient conditions and expresses their orientation counts through a transfer matrix, but its work record leaves the necessary local-limit estimate conditional. Qwen develops Gaussian/spanning-tree bounds and a multiscale argument covering $\log\log n=o(d^{1/3})$, yet cannot eliminate the residual size restriction. GLM-Flash's exact positive Eulerian-subgraph expansion supplies a useful reformulation without bounding the resulting partition function uniformly. DeepSeek-Flash explores Gaussian approximations, extremal reductions, and exact tableau counts for a structured family, while explicitly acknowledging that a uniform local-limit bound remains unavailable. Qwen rates the first three partial contributions nontrivial and DeepSeek-Flash's progress trivial. The differentiating step is selecting a representation whose error can be controlled uniformly in the unrestricted variable.

\subsection{ORB-PHYS-69: Sklyanin algebra classification in exceptional fields}
Extending the classification in \citet{iyudu2021sklyanin}, the task requires necessary and sufficient conditions over arbitrary fields in both regimes, including exceptional parameter families and the passage from an algebraic closure back to the original field (descent). Several models predict the same rule for the defining parameter triple $(p,q,r)$: equality up to a common nonzero scale, or exchange of the first two coordinates. The outcome difference lies in proving that no other isomorphisms occur.

Astra works from relation-space and tensor invariants, then handles the exceptional strata separately. Section 15 explains why a characteristic-zero cubic invariant loses information in characteristic three and replaces it with a contraction-determinant invariant. It also separates low-degree checks from the all-degree Hilbert-series argument, closing the latter using a central quotient rather than assuming regularity. GPT-5.6-Sol likewise derives field-independent invariants and explicitly covers characteristic two and exceptional descent cases. Both are judged solved by Qwen.

Qwen's solution combines a substitution-group description, descent, and exact finite-field enumeration of graded substitutions through transformed quadratic relation spaces. Nevertheless, its Sections 10.2--10.3 leave a special-line component hypothesis and characteristic-two proof unresolved. Qwen review therefore assigns breakthrough partial rather than solved. GLM, Kimi, GLM-Flash, and DeepSeek-Flash obtain useful generic-stratum, descent, or finite-field results but leave general completeness obligations. DeepSeek-Flash explicitly names the remaining ``no further isomorphisms'' assertion as Conjecture U. This case distinguishes evidence for a plausible classification from a proof covering every required field and stratum, highlighting the need to track quantifiers and unresolved dependencies.

\subsection{ORB-PHYS-82: Integrality of generalized lambda-function values}
The task extends results of \citet{ishii2011lambda} to excluded parameter regimes. It combines integral expansions after modular transformations and integrality over $\mathbb Z[j]$, where $j$ is the classical modular invariant, with corresponding consequences for complex-multiplication (CM) values. GLM-Flash, Astra, GPT-5.6-Sol, Kimi, GLM, Qwen, and DeepSeek-Flash are all judged solved by Qwen. This provides a shared success case for comparing the proof strategies of Flash and frontier models.

Their proof routes differ. GLM-Flash combines explicit cyclotomic valuation cases with symbolic elimination, deriving a triplication identity before removing the auxiliary variable. Qwen derives a Siegel-function product, reduces transformed expansions to leading coefficients, and classifies their valuations. DeepSeek-Flash uses a product over modular transforms and its prime valuations to obtain a norm obstruction to global integrality. The different successful routes show how distinct mathematical representations can support progress on the same open research problem.

\FloatBarrier
\section{Problem Coverage}
\begin{table}[h]
\centering\small
\caption{Primary-subfield composition. Each problem is assigned to one primary subfield for counting; the accompanying data give all assignments.}
\label{tab:coverage}
\begin{tabularx}{\linewidth}{@{}YrYr@{}}\toprule
Mathematics & $n$ & Physics & $n$ \\\midrule
Combinatorics and discrete mathematics & 12 & Statistical mechanics and integrable models & 21 \\
Geometry and topology & 8 & Algebraic/combinatorial mathematical physics & 7 \\
Algebraic geometry and commutative algebra & 7 & Quantum information and computation & 4 \\
Analysis, partial differential equations, and dynamics & 6 & Field theory, gauge theory, and strings & 3 \\
Number theory and arithmetic dynamics & 5 & Statistical optics & 1 \\
Representation theory and operator algebras & 4 & & \\
Optimization and algorithms & 3 & & \\
Set theory and logic & 1 & & \\\midrule
Total & 46 & Total & 36 \\\bottomrule
\end{tabularx}
\end{table}
{\small
\begin{longtable}{@{}p{.20\linewidth}p{.76\linewidth}@{}}
\caption{The 82 retained problem statements. Titles are inherited from the archive. Domain prefixes record collection provenance.}\\
\toprule ID & Problem \\\midrule\endfirsthead
\toprule ID & Problem (continued) \\\midrule\endhead
\bottomrule\endfoot
ORB-MATH-01 & Irrationality and arithmetic nature of the Euler--Mascheroni constant \\
ORB-MATH-02 & Griffiths' 1969 positive-polynomial problem at the level of Chern forms: pointwise (weak) positivity of Schur forms for Griffiths-positive vector bundles \\
ORB-MATH-03 & Algebraicity of Weil classes on polarized abelian varieties of Weil type \\
ORB-MATH-04 & Quillen's conjecture $Q_n$: are all projective modules over the localization of a regular local ring at a regular parameter free? \\
ORB-MATH-05 & Binary Goldbach conjecture: is every even integer greater than 2 a sum of two primes? \\
ORB-MATH-06 & The Gorenstein bimodule conjecture: is a finite-dimensional algebra selfinjective exactly when its regular bimodule is Gorenstein projective? \\
ORB-MATH-07 & Iitaka's subadditivity conjecture for the logarithmic Kodaira dimension \\
ORB-MATH-08 & The Tate conjecture for divisors on surfaces over a finite field in Kahn's affine-open reformulation \\
ORB-MATH-09 & The motivic t-structure conjecture: existence of a non-degenerate tensor- and realization-compatible t-structure on Voevodsky's category of geometric motives \\
ORB-MATH-10 & Arithmetic nature of the odd Riemann zeta values: irrationality of $\zeta(2m+1)$ for $m\geq 2$ \\
ORB-MATH-11 & The Gorenstein Symmetric Conjecture for Artin Algebras \\
ORB-MATH-12 & The Bost conjecture: is the $\ell^1$-assembly map an isomorphism for every countable discrete group? \\
ORB-MATH-13 & Does every connected Cayley graph on a finite group have a Hamiltonian cycle? (The Cayley graph Hamiltonicity conjecture) \\
ORB-MATH-14 & Unconditional uniform boundedness of preperiodic points for unicritical polynomials $z^d+c$ over number fields \\
ORB-MATH-15 & The 11/8-conjecture (Matsumoto): $b_2\ge\frac{11}{8}|\sigma|$ for closed smooth spin 4-manifolds \\
ORB-MATH-16 & The generalized Chern conjecture: is the Euler characteristic an obstruction to flat structures on closed aspherical manifolds? \\
ORB-MATH-17 & The optimal universal constant in linear lower bounds for the crossing number of composite knots \\
ORB-MATH-18 & The planar embedding conjecture: constant-distortion embeddings of planar graph metrics into $L_1$ \\
ORB-MATH-19 & The Collatz (3n+1) conjecture, equivalently the nonvanishing of the arrival coefficients in Masetti's iterative functional equation \\
ORB-MATH-20 & Bilinear Bochner--Riesz means: the $\delta=0$ bilinear disc multiplier in one dimension and the $L^2\times L^\infty\to L^2$ smoothness threshold in higher dimensions \\
ORB-MATH-21 & Exact value of the univalent Bloch--Landau constant $U$ \\
ORB-MATH-22 & The Arithmetic Persistence Conjecture: does arithmetic hyperbolicity persist under extension of the base field? \\
ORB-MATH-23 & Are all star flows multisingular hyperbolic? The full characterization of star flows beyond dimension 3 \\
ORB-MATH-24 & The Weinstein conjecture for closed contact manifolds of dimension at least five \\
ORB-MATH-25 & The strong Weinstein conjecture: existence of null-homologous Reeb links on closed contact manifolds \\
ORB-MATH-26 & One-dimensionality of stable Allen-Cahn solutions with bounded energy density in dimensions 5 through 7 \\
ORB-MATH-27 & Existence of global energy-admissible weak solutions of the incompressible Euler equations for arbitrary initial data in the energy space \\
ORB-MATH-28 & The Penrose inequality for general (non-time-symmetric) asymptotically flat initial data sets \\
ORB-MATH-29 & The Kirchhoff--Hopf problem: does the six-sphere admit a complex structure? \\
ORB-MATH-30 & Parametrization of the volume conjecture for hyperbolic knots beyond the figure-eight knot \\
ORB-MATH-31 & Is Axiom A dense in the $C^1$ topology among diffeomorphisms of compact surfaces? (Smale's problem / $C^1$ Palis density conjecture, surface case) \\
ORB-MATH-32 & Existence of finite projective planes of non-prime-power order (the prime power conjecture) \\
ORB-MATH-33 & Sawin's coupling question for the entropy approach to Frankl's union-closed sets conjecture \\
ORB-MATH-34 & Graph-restrictiveness of semiprimitive groups with a regular normal nilpotent subgroup: the exceptional cases where 2 or 3 divides the degree (the Poto\v{c}nik--Spiga--Verret conjecture) \\
ORB-MATH-35 & Is every line graph $\{1\star(\chi'(G)-3),\,3\}$-choosable? \\
ORB-MATH-36 & Growth rate of the higher-dimensional Erd\H{o}s--Szekeres function $ES_d(n)$ for fixed $d\ge 3$ (Füredi's conjecture) \\
ORB-MATH-37 & The strong Atiyah conjecture for groups with bounded torsion \\
ORB-MATH-38 & Ryser's conjecture: full transversals in Latin squares of odd order \\
ORB-MATH-39 & Clustered colouring of graphs excluding a fixed minor: the Norin--Scott--Seymour--Wood conjecture $\chi_\star(\mathcal{M}_H)\le 2\,\overline{td}(H)-2$ \\
ORB-MATH-40 & Seymour's Second Neighborhood Conjecture \\
ORB-MATH-41 & Bouchet's 6-flow conjecture for flow-admissible signed graphs \\
ORB-MATH-42 & Tutte's 3-Flow Conjecture \\
ORB-MATH-43 & Existence of a polynomial pivot rule for the simplex method (the combinatorial core of Smale's Problem 9) \\
ORB-MATH-44 & The 4/3-Conjecture for the Integrality Gap of the Subtour (Held--Karp) Relaxation of the Metric Traveling Salesman Problem \\
ORB-MATH-45 & Schalekamp--Williamson--van Zuylen conjecture: is the integrality gap of the subtour LP attained on half-integral vertices? \\
ORB-MATH-46 & Woodin's HOD Conjecture: can the HOD Hypothesis fail in the presence of large cardinals? \\
ORB-PHYS-47 & Rigorous, assumption-free equivalence of the two contour prescriptions in superstring perturbation theory (low spacetime dimensions, stub-free intermediate-state sums, and controlled integral interchanges) \\
ORB-PHYS-48 & Coexistence of infinite $+$ and $-$ clusters in the Ising model on the square matching lattice $\mathbb{Z}_*^2$ for all subcritical temperatures \\
ORB-PHYS-49 & Symmetry algebra underlying the transfer-matrix degeneracies of the eight-vertex model at elliptic roots of unity \\
ORB-PHYS-50 & Finite non-Abelian gluing variety and the classification of directly empirically significant regional gauge symmetries in Yang--Mills theory \\
ORB-PHYS-51 & A fault-tolerance threshold theorem for adiabatic quantum computation under simultaneous decoherence and control errors \\
ORB-PHYS-52 & Almost-everywhere classical simulation of quantum query algorithms: the quantum analogue of average-case derandomization of query algorithms (Aaronson--Ambainis simulation conjecture) \\
ORB-PHYS-53 & Is the quantum capacity of a general quantum channel uncomputable? \\
ORB-PHYS-54 & Scheme-independent quark-gluon decomposition of the nucleon mass in QCD and the physical role of the trace anomaly \\
ORB-PHYS-55 & Classification of planar stochastic electromagnetic sources whose beams have propagation-invariant degree of polarization \\
ORB-PHYS-56 & Exact fixed-magnetization partition functions and Helmholtz fluctuation-induced forces beyond one dimension and beyond the Gaussian fixed point \\
ORB-PHYS-57 & Exact critical manifold of the five-probability inhomogeneous bow-tie lattice: full-domain proof and rigorous foundation of the negative-probability balancing construction \\
ORB-PHYS-58 & Exact solution and exact critical temperature of the spin-one Ising model on the honeycomb lattice \\
ORB-PHYS-59 & Complete classification of maximum-velocity quantum circuits: the two-qubit conjecture and the general local-dimension problem \\
ORB-PHYS-60 & Constructive derivation of the monomer--topological-defect correspondence from the Grassmann-integral representation of the two-dimensional classical dimer model \\
ORB-PHYS-61 & Explicit identification of the local and Fourier inverse Kasteleyn matrices for the critical Z-invariant Ising model on periodic isoradial graphs \\
ORB-PHYS-62 & Universal validity of Pauling's residual entropy estimate for regular graphs with growing degree \\
ORB-PHYS-63 & Edge fluctuations of the melting domain wall in the XXZ spin-1/2 chain at nonzero anisotropy: the Saenz--Tracy--Widom conjecture \\
ORB-PHYS-64 & Griffiths--Kelly--Sherman-type correlation inequalities for the eight-vertex model with non-uniform vertex weights \\
ORB-PHYS-65 & Analytic evaluation of the square-lattice k-coloring entropy W(k) beyond the exactly solved k=3 ice point: higher-cycle algebra, non-Eulerian subgraphs for k>3, and finite-width Toeplitz transfer-matrix recursions \\
ORB-PHYS-66 & Explicit closed-form evaluation of the partial partition functions of the three-coloring model with domain wall boundary conditions in the general elliptic case \\
ORB-PHYS-67 & Proof of the inhomogeneous eigenvalue conjecture for the supersymmetric eight-vertex model on a strip with open boundary conditions (Hagendorf--Liénardy Conjecture 5.1) \\
ORB-PHYS-68 & Universality of the critical finite-size free-energy correction with central charge $c=1$ in non-solvable Ashkin--Teller and eight-vertex models \\
ORB-PHYS-69 & Isomorphism classification of three-generator Sklyanin algebras over fields of characteristic 3 and over fields without a nontrivial cubic root of unity \\
ORB-PHYS-70 & Proof of the Kotousov--Lukyanov conjectures for the scaling-limit norms of Bethe states in the critical XXZ Heisenberg chain \\
ORB-PHYS-71 & Iterated binary fusion versus single m-fold fusion for families of modules over the enlarged periodic Temperley--Lieb algebra \\
ORB-PHYS-72 & Rigorous proof of the universal MPS--Bethe-state overlap formula for orthogonal and symplectic rational spin chains \\
ORB-PHYS-73 & Exact critical coupling of the q-state Potts model on the kagome lattice \\
ORB-PHYS-74 & Rigorous justification of the replica coordinate Bethe ansatz for the log-gamma polymer: Bethe-string norms, completeness, and analytic continuation in replica number \\
ORB-PHYS-75 & Analytic origin of the $1/\log(LM)$ finite-size correction to the four-state Potts spin-cluster crossing probability \\
ORB-PHYS-76 & Deletion--contraction recurrences for local flows and tensions on non-orientable surfaces \\
ORB-PHYS-77 & Completing the approximability classification of the parameterized Tutte polynomial at rational points with $(x-1)(y-1)\notin[0,1]$ \\
ORB-PHYS-78 & Arbitrary-width structure of the Potts partition-function-zero accumulation locus on self-dual square-lattice strips (the Chang--Shrock conjecture) \\
ORB-PHYS-79 & Pair-correlation functions of the two-dimensional integrable N-state chiral Potts model \\
ORB-PHYS-80 & Fermionic sum sides for $\mathcal{W}_r$ minimal-model characters and $r$-cylindric partition generating functions at ranks $r\geq 4$ \\
ORB-PHYS-81 & Explicit fermionic-type character formulas for Feigin--Stoyanovsky type subspaces of standard $\mathfrak{sl}(\ell+1,\mathbb{C})^{\sim}$-modules in rank $\ell\geq 3$ \\
ORB-PHYS-82 & Integrality of singular values of the generalized lambda function $\Lambda_k=W_{[k,2,1]}$ in the excluded regime $\delta=\gcd(k,N)>1$ with $3\mid\delta$ or $N/\delta$ a prime power \\
\end{longtable}}
\end{document}